\documentclass[runningheads]{llncs}
\usepackage[hidelinks]{hyperref}
\usepackage{subfigure}
\usepackage{algorithm}
\usepackage{algpseudocode}
\usepackage{graphicx}
\usepackage{amsmath,amssymb,amsfonts}
\usepackage{multirow}
\usepackage{tabularx,booktabs}
\usepackage{xfrac}
\usepackage{lipsum}

\usepackage{multirow}
\usepackage{rotating}
\usepackage{makecell}
\usepackage{booktabs}

\begin{document}
\title{Transfer Learning for Instance Segmentation of Waste Bottles using Mask R-CNN Algorithm}


\titlerunning{Instance Segmentation of Waste Bottles Using Deep Learning}

\author{Punitha Jaikumar\inst{1} \and Remy Vandaele\inst{1,2} \and Varun Ojha\inst{1}}

\authorrunning{Jaikumar, et al.}
%

%
\institute{Department of Computer Science, University of Reading, Reading, UK  \and Department of Meteorology, University of Reading, Reading, UK \\    
\email{\{p.jaikumar,r.a.vandaele,v.k.ojha\}@reading.ac.uk}\\
}

%
\maketitle              
\begin{abstract}
This paper\footnote[1]{cite as: Jaikumar, P., Vandaele, R., Ojha, V. (2021). Transfer Learning for Instance Segmentation of Waste Bottles Using Mask R-CNN Algorithm. Intelligent Systems Design and Applications. ISDA 2020. Springer.} proposes a methodological approach with a transfer learning scheme for plastic waste bottle detection and instance segmentation using the \textit{mask region proposal convolutional neural network} (Mask R-CNN). Plastic bottles constitute one of the major pollutants posing a serious threat to the environment both in oceans and on land. The automated identification and segregation of bottles can facilitate plastic waste recycling. We prepare a custom-made dataset of 192 bottle images with pixel-by pixel-polygon annotation for the automatic segmentation task. The proposed transfer learning scheme makes use of a Mask R-CNN model pre-trained on Microsoft COCO dataset. We present a comprehensive scheme for fine-tuning the base pre-trained Mask-RCNN model on our custom dataset. Our final fine-tuned model has achieved 59.4 \textit{mean average precision} (mAP), that corresponds to the MS COCO metric. The results indicate promising application of deep leaning for detecting waste bottles.

\end{abstract}
\section{Introduction}
Over the years, the plastic bottle has evolved from being one of the most useful materials available to a cause for the severe problem of pollution on land and seas~\cite{jambeck2015plastic}. Plastic waste poses a serious threat to the environment. The Ocean Conservancy annual report mentions that plastic bottles ranked among the top~5 pollutants~\cite{ocean2019}.  In 2019, Ocean Conservancy reportedly collected 1.75~million bottles from the beach cleanup initiative~\cite{ocean2019}. In order to alleviate the problem, organizations around the world have taken various measures to collect and segregate the pollutants for plastic waste recycling and their eco-friendly disposal. Among those measures, drones and video surveillance in urban waste management along with \textit{image processing systems} can be used to alert environmental agencies to take adequate measures.

In this work, we study the performance of a \textit{deep learning} algorithm for detecting the waste bottles from a set of images. We prepared a custom dataset of 192 images for the training and validation of our deep learning bottle detection model developed on \textit{mask region proposal convolutional neural network} (Mask R-CNN). Since object shapes and forms can vary, a dataset of 192 images may not constitute an adequate training set. Therefore, we developed a methodological framework using transfer learning~\cite{pan2009survey} to exploit a model pre-trained with over 5000 images 'minival' subset of the Microsoft Common Object in Context (COCO) dataset pertaining to 'val2014training' images of 80 object categories~\cite{lin2014microsoft}. Further, the limitation of the small custom-made dataset is addressed by applying data augmentation techniques during the fine-tuning process to improve the model's performance~\cite{ImgAugsurvey2019}. 

Our fine-tuned model was trained on the custom-made dataset and when evaluated on instance segmentation offered a 59.4 \textit{mean average precision} (mAP) following the COCO evaluation metric, \textit{intersection over union} (IoU) of threshold range [0.5:0.05:0.95]. Also, the final fine-tuned Mask R-CNN model was able to detect bottles from test images and videos. The trained models, the outputs (images and videos), and the relevant Mask R-CNN open-source code (adapted from~\cite{abdulla2017mask}) is available at repository~\cite{punitha2020}.

This paper is organized as follows: Sec.~\ref{sec:related_work} refers to the relevant literature. Sec.~\ref{sec:Implement} describes the instance segmentation model training methods. The obtained results are described in Sec.~\ref{sec:results}, followed by conclusions in Sec.~\ref{sec:con}.

\section{Related Work}
\label{sec:related_work}
\begin{figure}[h!]
    \centering
    \includegraphics[height=4cm,width=12cm]{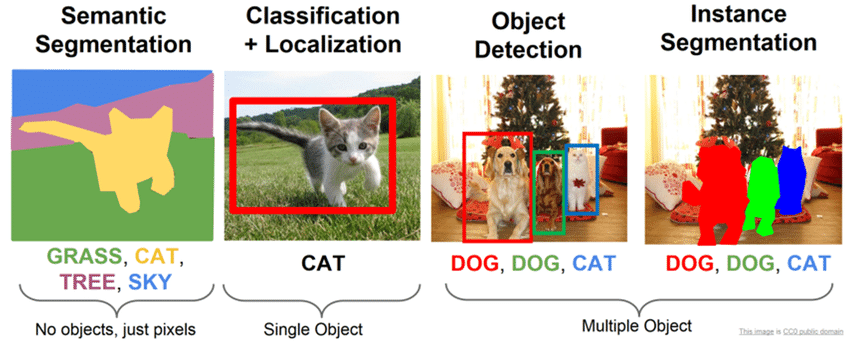}
    \caption{Comparison of semantic segmentation, classification and localization, object detection and instance segmentation. (Figure adapted from~\cite{det2017})}
    \label{fig:seg}
    
\end{figure}

The object segmentation task has evolved over the years in the field of computer vision. Image-based systems have revealed the potential to find an automated solution for the environmental problems and research is initiated for application in waste segregation for recycling purpose. The machine learning techniques such as support vector machine, k-nearest neighbor, decision tree, and logistic regression were used for plastic bottles waste management related work~\cite{kambam2019classification,kokoulin2018convolutional,yang2016classification}. The recent advancements in deep neural networks have outperformed traditional machine learning models in the object classification and detection tasks. Deep Learning in the field of computer vision developed from image classification to object localization, object segmentation, and to instance segmentation~\cite{garcia2018survey} (cf. Fig.~\ref{fig:seg}). There are two type of object detectors: two stage detectors-Region based R-CNN family and single stage detectors: such as \textit{you look only once} (YOLO)~\cite{redmon2016yolo} and  \textit{single shot detector} (SSD)~\cite{Liu_SSD2016}.

\textbf{Two-Stage Detectors}
The \textit{region proposed} methods of object detection are from the R-CNN family. The model pipeline detects objects in two stages: (1) generating proposal of \textit{region of interest} (RoI) and (2) classification of objects in the proposed regions. The \textit{region proposal} algorithms have evolved in the following order~\cite{zhao2019objectreview}: R-CNN, Fast R-CNN, Faster R-CNN and Mask R-CNN.

\textbf{Single-Stage Detectors}
The single-stage detectors take an approach of a simple regression problem predicting the bounding box coordinates and the class probabilities from images in one evaluation. The YOLO~\cite{redmon2016yolo} and SSD~\cite{Liu_SSD2016} families belong to this category of detector, and are designed for speed and real-time use while they compromise on accuracy.

In~\cite{zhao2019objectreview}, authors discussed the approaches adopted in the object segmentation and compares the performance of various algorithms and its benchmark. The training time versus accuracy will be the main practical trade-off for detector selection when compared to other image segmentation and object detection algorithms and methods~\cite{huang2017speed}. Two-stage region proposed detectors performed better over single stage detectors. In~\cite{fulton2019robotic}, the author concluded that Faster R-CNN has significantly outperformed models when applied on the dataset containing plastic bottle images of underwater debris. In~\cite{wang2018bottle}, authors used Faster R-CNN, \textit{single-shot multi-box detector} (SSD), and YOLO version 2 (YOLOv2) architectures; and compared their performances. They show that the Faster R-CNN with the rotational region proposed network achieved 90.3\% PASCAL VOC AP precision~\cite{everingham2015pascal}, followed by 90.1\% achieved by SSD and 77.4\% achieved by YOLOv2.  

\begin{figure}[h!]
    \centering
    \includegraphics[width=0.98\textwidth]{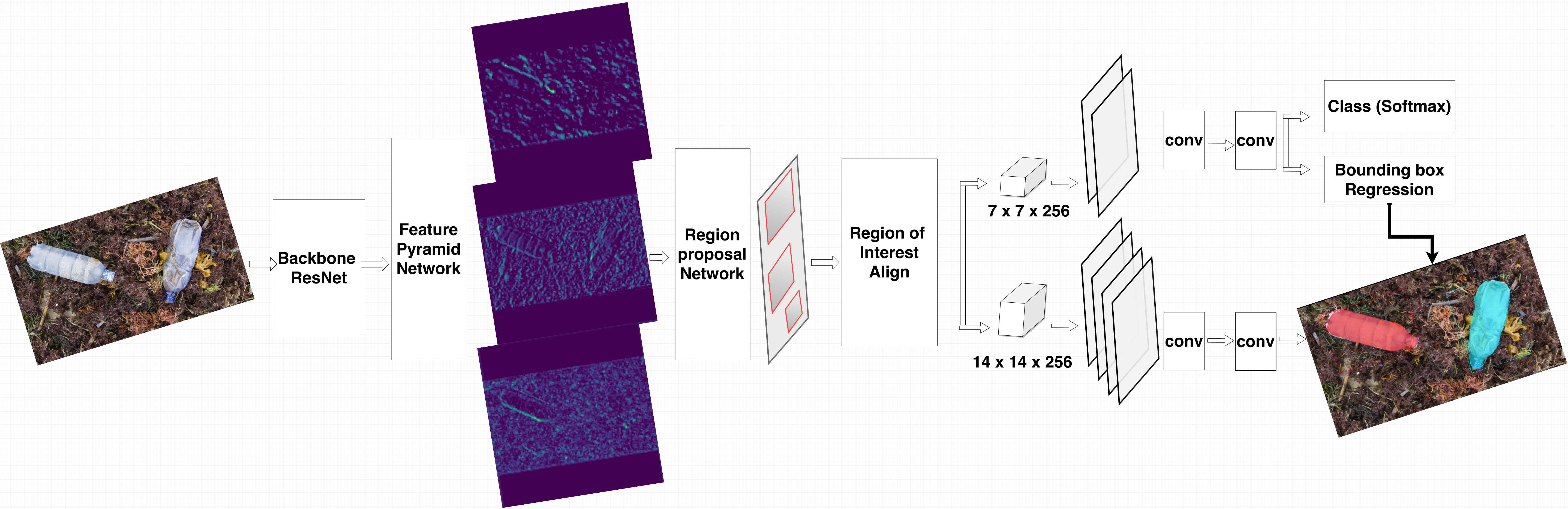}
    \caption{Mask R-CNN}
    \label{fig:Maskhead}
\end{figure}

\textbf{Mask R-CNN:} This algorithm belongs to the two-stage detectors that uses \textit{feature pyramid network} (FPN), and \textit{region proposal network} (RPN) for the object segmentation. It is an extension of the Faster R-CNN~\cite{ren2015faster} along with a new pipeline for masking the detected objects for instance segmentation~\cite{he2017mask} (cf. Fig.~\ref{fig:Maskhead}). In Mask R-CNN, the spatial layout of input object encoded by a mask and predicted by using fully connected network (FCN) pixel to pixel convolutions instead of the fully connected (FC) layers that will flatten into vectors lacking spatial dimensions as in Faster R-CNN. The RoI-align (instead of RoI-pooling) calculated with bi-linear interpolation to improve segmentation accuracy. This technique had gained popularity as it improves object instance segmentation precision~\cite{garcia2018survey,huang2017speed,zhao2019objectreview}. In this work, for instance segmentation of bottles at pixel-level, we opted a two-stage detector: Mask R-CNN.

\section{Bottle Segmentation using Mask R-CNN} 
\label{sec:Implement} 
The implementation of our bottle segmentation model started with the datasets preparation. The Mask R-CNN model was built with Python 3, TensorFlow, Keras, and OpenCV libraries~\cite{chollet2015others} by adapting the code from open-source Mask R-CNN implementation~\cite{abdulla2017mask}. The annotated custom dataset along with pre-trained model weights were fed into the training pipeline consisting of several stages.

\textbf{Model configuration.} The ResNet-50 and ResNet-101 were the \textit{feature backbone network} options for Mask R-CNN algorithm. The images were resized to dimension $1024\times1024$. The initial RoI parameter was 200 for training of the classifier/mask heads. Later, for the incremental training, RoI parameter was 512 and the maximum ground truth instances for one image was updated from 100 to 512. Also, for incremental training, the \textit{detection maximum instance} was updated from 150 to 512 that facilitates the maximum detection of object instances. The \textit{detection minimum confidence} parameter values were 0.7 and 0.5 for tuning model's performance. The parameter setup and training process instructions are provided in~\cite{punitha2020}.

\textbf{Dataset description.} The dataset consists of custom images downloaded from the internet for this work. Though there are many state-of-the-art datasets such as COCO and the PASCAL VOC that have general images for training and research purpose, it was noted that the availability of images pertaining to the waste bottle class was limited. To overcome this problem, a discrete set of images were downloaded from the internet. The set contains single and multiple bottle images of normal and deformed features possessing various backgrounds (cf. Table~\ref{tab:dataset}). Detecting objects in multi-bottle images would be a challenge due to occlusion and lighting conditions. This will be a challenge for the generalization of the bottle instance segmentation for the model. 

\textbf{Data pre-processing.} Image annotation of each object is a time-consuming and tedious task. However, it is an essential initial step for an instance segmentation task. It is necessary to define the  pixel-wise ground truth for the target objects for the models to perform segmentation and mask generation. Unlike other object detectors, the Mask R-CNN model requires pixel-wise annotation for training. There are many publicly available tools for data annotation such as VIA-VGG\cite{vgg} and LabelMe \cite{torralba2010labelme}. 

For this work, pixel-wise polygon annotation of the images was performed for instance segmentation training as per the COCO dataset format. Custom dataset images were of varied sizes and formats. VIA-VGG annotator tool  was used for the pixel-wise polygon graphical annotation. The tool generates the output in json format for the annotated images. The segmented ground truth mask represents a region-wise \textit{spatial position} and \textit{axis} of each target object. The annotated dataset also includes two \textit{no-instance} images and two \textit{partial-annotation} images. The images were resized to dimension $1024\times1024\times3$ and to preserve the \textit{aspect ratio}, each image was padded with zero to match the one-size training requirements (square format of same dimension). The dimensions of the images range from small images of $ 150\times255$ to $2448\times3264$ in the dataset.
All the images stored in the system in two folders with the training images and the pixel annotation and labeling information as arrays for each image in the json file. These image files are read one-by-one from the system drive for resizing. Video files input for testing is pre-processed with OpenCV libraries in python to extract the images frame-by-frame to be fed into the detection pipeline for segmentation. The segmented output image frames were saved into the set directory path with the OpenCV video.

\begin{table}
    \centering
    \caption{Custom dataset of bottles downloaded from Internet}
    \label{tab:dataset}
    \begin{tabular}{p{1cm} p{1.3cm} p{1.3cm} p{1.3cm} p{1.3cm} p{1.5cm} p{1.3cm} p{2cm}}
        \bottomrule
        Type  & No. of Images & Waste Bottle Specif Images & Total Bottle Instance Count & Bulk Bottle Instance Count & Image Range-min pixel & Image Range-max pixel & Images Resized (RGB)\\
        \midrule
        Train & 177 & 102 & 712 & 561 & 33750 & 7990272 & $1024\times1024\times3$ \\
        Val & 15 & 3 & 113 & 104 & 535824 & 2250000 & $1024\times1024\times3$\\
        \bottomrule
    \end{tabular}
\end{table}

\textbf{Transfer learning scheme.}  The custom dataset training was initialized with the Mask R-CNN model pre-trained on MS-COCO dataset with 80 classes. The model weights in the initial layers representing low-level features will be useful in many classification tasks. While deeper layers learn the high-level features, this can be altered and retrained according to the problem definition. In the initial phase, the models were trained with backbone ResNet-50 and ResNet-101 for the waste bottle instance segmentation using \textit{stochastic gradient descent} (SGD) and \textit{Adam} optimizer. The deeper network of ResNet-101 and SGD optimizer performed better comparatively and was opted for further incremental training experiments on the custom dataset. Our transfer learning scheme has the following stages for pre-trained model fine-tuning:
\begin{itemize}
    \item Stage 1: Head-layer training with all other layers frozen
    \item Stage 2: Fine-tuning selected layer or all layers
    \item Stage 3: Extended training (optional)
\end{itemize}

Table \ref{tab:Phase2param} shows the detailed experiment for the transfer learning scheme adopted to train for bottle segmentation. \textbf{Stage 1:} \textit{Head layers} trained with Data Augmentation for 30 epochs initiated for ResNet101 Backbone. \textbf{Stage 2:} As part of model fine-tuning, the {stage 2} models trained with the learned weights of models in the previous stage.
The \textit{4+ layers} and \textit{ALL layers} models trained for 30 epoch with selected data augmentation. Additionally, the \textit{4+ layers} models M4 and M10 trained up to 100 epochs. And, as part of \textbf{Stage 3}, models trained for an additional 20 epochs to assess whether it improves the performance. 
With respect to the \textit{ALL layers} training at {stage 3}, the \textit{4+ layers} trained models M4 and M10 tuned up to 30 epochs initially and then up to 100 epochs. Incrementally, in {stage 3}, models were trained up to 150 and 160 epochs for \textit{4+ layers} and \textit{ALL layers} respectively. Each model after training were evaluated by using COCO evaluation metric and compared against the model loss.

\textbf{Model evaluation and optimization.} Models performances were compared against loss metrics and were evaluated by using COCO evaluation metric mAP~\cite{he2017mask,lin2014microsoft}. Image segmentation models being far too expensive for the cross-validation method, Mask R-CNN hyper-parameter tuning was based on the configuration parameters.  

\section{Results and Discussion}


Table~\ref{tab:Phase2M1to17Eval} shows the models \textit{mean average precision} (mAP) results as per the MS COCO evaluation metrics- AP$^{50}$, AP$^{75}$, AP$^{95}$ and mAP[0.5:0.95:05]. The {stage 2} model M11 performed well compared to all other models trained using the transfer learning scheme. The \textit{4+ layers} with augmentation model in {stage 2} at the 100-th epoch achieved 53.0 mAP and at {stage 3} tuning of 20 epochs shows gradual improvement achieving 55.3 mAP for both \textit{ALL layers} and \textit{4+ layers}. Tuned model for \textit{ALL layers}  without augmentation performed better in {stage 2}, achieving mAP for model M3 and in stage 3, achieving 56.2 and 56.4 mAP for models M12 and M14 respectively. AP$^{90}$ increased from 13.90 precision in stage 1 to 49.20 precision in {stage 3} fine tuning.

Models trained without data augmentation at {stage 2} and {stage 3} performed better comparatively. This could be due to the model learning with fine-tuned feature with augmentation having less diversity in the training data as it was already trained with augmentation in \textit{Head layer} training. In this task, only horizontal left-right flip was used, more varied augmentation if applied could have shown improvement in the model performance at later stages of tuning.

Further, model behavior requires to be monitored at each incremental training as the false-positives in-addition to true positives may affect the quality of the output. The model's performance did not improve much in later stages of tuning, this could be attributed to an inadequate dataset. While training \textit{ALL layers}, it is important to decide the training based on the dataset as most of the low-level features will be changed drastically; and it can impact the output on extended training leading to over-fitting. 

\begin{figure*}[h!]
\centering
  \begin{subfigure}
    {\includegraphics[height=2.5cm,width=0.32\linewidth]{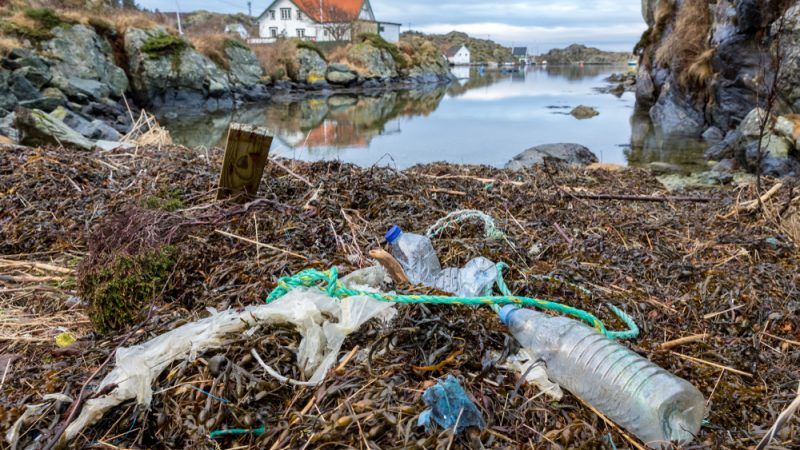}}
    {\includegraphics[height=2.5cm,width=0.32\linewidth]{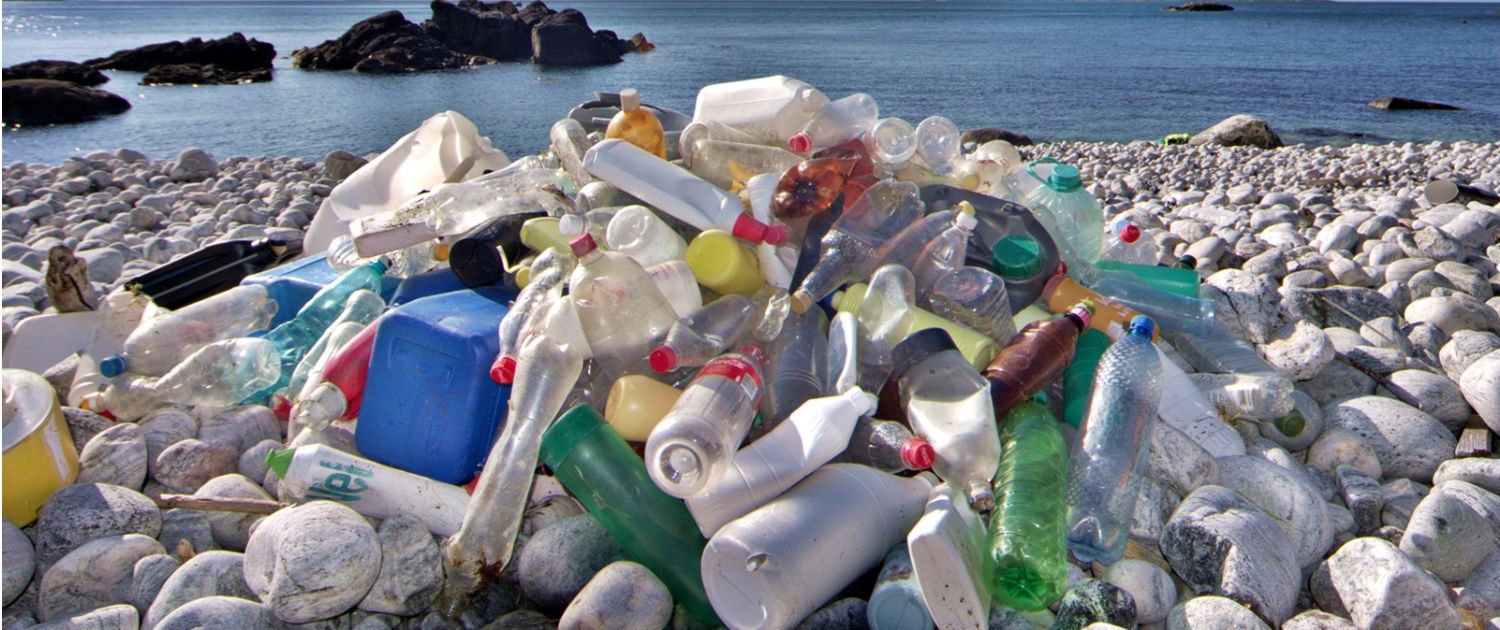}}
    {\includegraphics[height=2.5cm,width=0.3\linewidth]{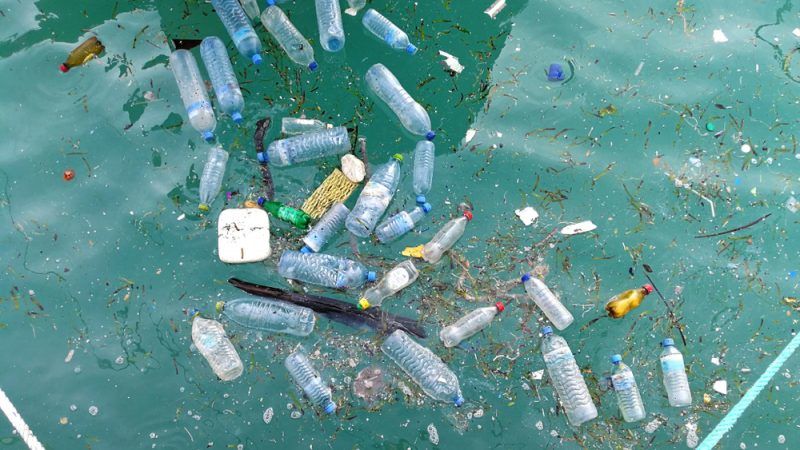}}
    \end{subfigure}
  
  \begin{subfigure}
    {\includegraphics[height=2.5cm, width=0.33\linewidth]{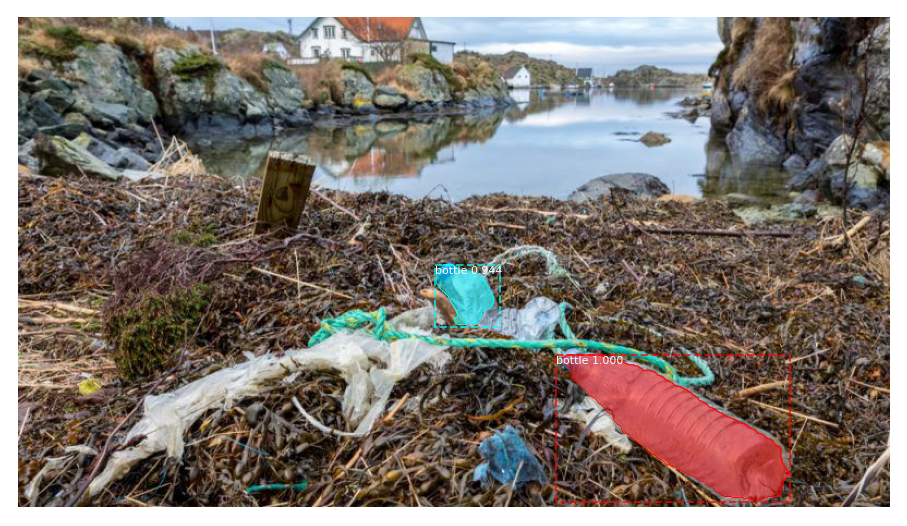}}
    {\includegraphics[height=2.5cm, width=0.32\linewidth]{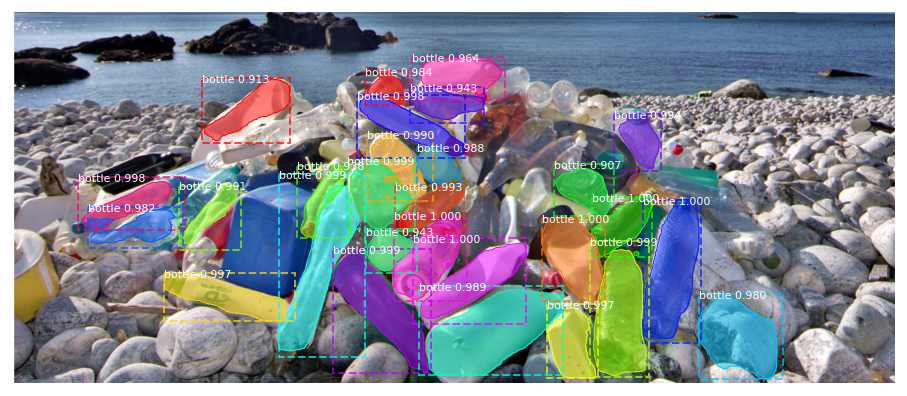}}
    {\includegraphics[height=2.5cm, width=0.31\linewidth]{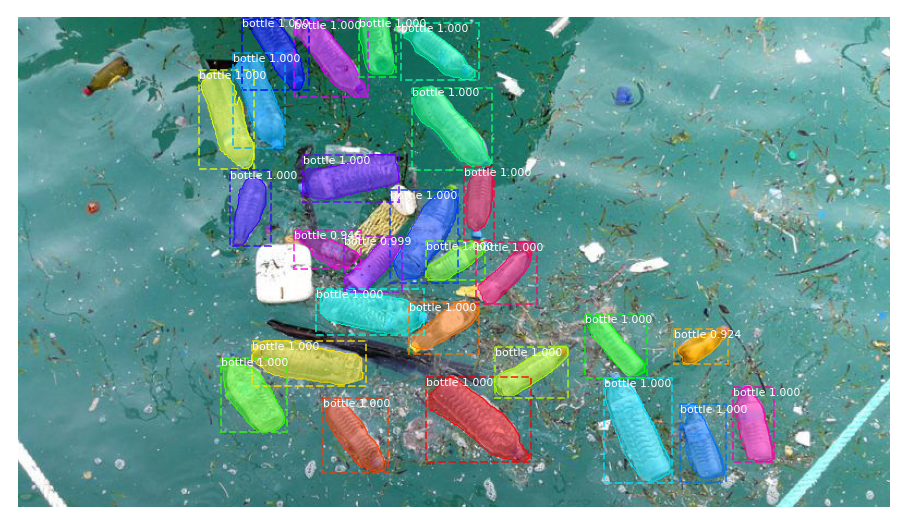}}
    \caption{Test image segmentation results}
    \label{fig:test}
  \end{subfigure}
\end{figure*}

\begin{figure*}[h!]
    \begin{subfigure}
    [90\% Detection 
    minimum\_confidence]{\includegraphics[height=2.5cm,width=0.3\textwidth]{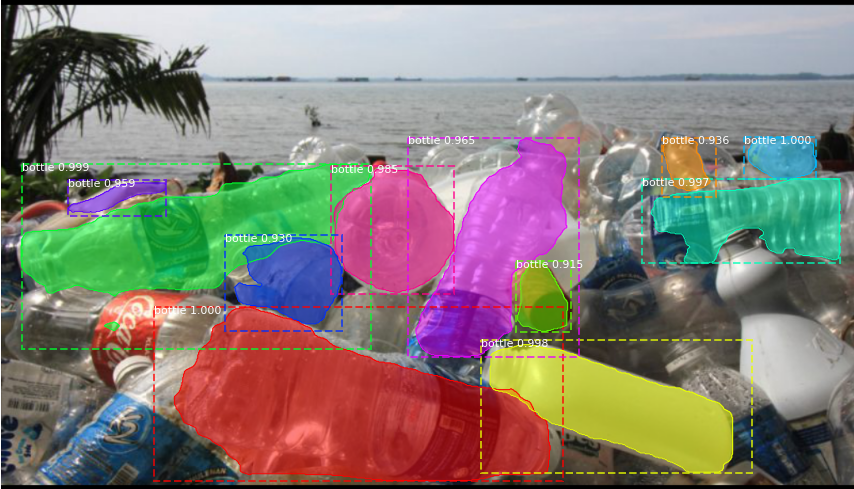}}
    \label{fig:f7}
    \end{subfigure}
    \hfill
    \begin{subfigure}
    [70\% Detection 
    minimum Confidence]{\includegraphics[height=2.5cm,width=0.3\textwidth]{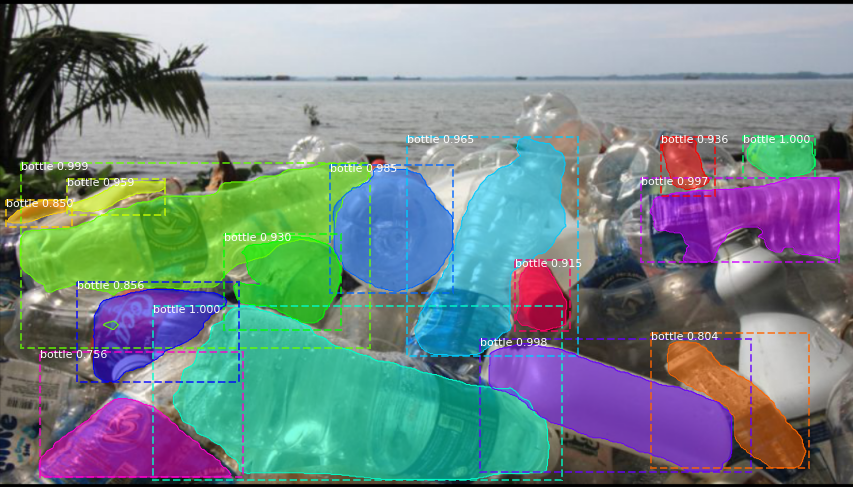}}
    \label{fig:f8}
    \end{subfigure}
    \hfill
    \begin{subfigure}
    [50\% Detection 
    minimum Confidence]{\includegraphics[height=2.5cm,width=0.3\textwidth]{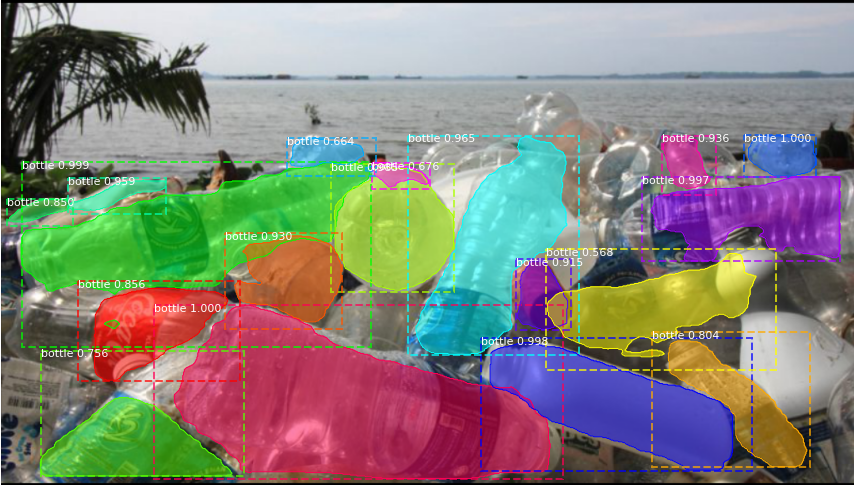}}
    \label{fig:f9}
    \end{subfigure}
    \caption{Dense instance segmentation}
    \label{fig:dense}
\end{figure*}

The model M11 achieved 59.4 mAP and 74.6 precision for AP$^{50}$ at stage 2 fine tuning for instance segmentation of bottle images including dense overlapped instances of varied shapes and features compared to the study done with Faster R-CNN models that shows detection of objects (including other class objects) with no overlap in \cite{fulton2019robotic} with 60.6 mAP and  in \cite{wang2018bottle} with 86.4 precision for AP$^{50}$ and 90.3 for rotational RPN.
Figs.~\ref{fig:test} shows the sample test image segmentation for the model M11 and Fig.~\ref{fig:dense} shows the instance segmentation of dense agglomerate of bottles. The number of instances segmented by the model increases with the decrease in the detection minimum confidence level.

\label{sec:results}
\begin{table}[h!]
    \scriptsize{
    \centering
    \caption{Transfer learning(TL) scheme: incremental step by step fine-tuning.}
    \label{tab:Phase2param}
    
    \begin{tabular}{p{1.2cm} p{2.1cm} p{1.5cm} p{1.5cm} p{2cm} p{1.2cm} p{1.2cm}}
        \toprule
    Model Ref No. & Starting Weights$^a$& TL-Stage & Training Layers & Augmentation & epochs & Total epochs\\
    \midrule
   \textbf{M1}   & COCO$^*$ &1& HEADS  &  Y & 30 & 30 \\&\\
 
    M2  & {M1 (30.h5)} &  2  & ALL &  Y & 30 & 60 \\
    M3  & {M1 (30.h5)} &  2  & ALL & N & 30 & 60\\&\\
    
   \textbf{M4}  & {M1 (30.h5)} & 2  & 4+ & Y & 30 & 60  \\
    {M5}  & \textbf{M4} (30.h5) &  2  & 4+ & Y & \textbf{70}  & 130  \\
    M6  & M5 (100.h5) &  \textbf3  & ALL & Y & 20 & 150 \\
    M7 & M5 (100.h5) &  \textbf3  & 4+ & Y & 20 & 150\\&\\
    
    {M8}     & \textbf{M4} (30.h5) &  3  & ALL & Y & 30 & 90    \\
    M9     & M8 (30.h5)   &  3  & ALL &  Y & 70   & 160\\ &\\
    
    \textbf{M10}    & {{M1} (30.h5)} &  2  & 4+ &  N & 30 & 60     \\
    {M11}     & {\textbf{M10} (30.h5)}&  2  & 4+ & N & \textbf{70}  & 130   \\
    M12   & {M11 (100.h5)}  &  \textbf3  & ALL & N & 20 & 150   \\
    M13   & {M11 (100.h5)}   &  \textbf3  & 4+ & N & 20 & 150\\ &\\

    {M14}   & {\textbf{M10} (30.h5)} &  3  & ALL & N & 30 & 90     \\
    M15   & {{M14} (30.h5)}   &  3  & ALL & N & 70   & 160\\
    \bottomrule
    \end{tabular}}
    {\scriptsize
    $^*$COCO dataset-Pre-trained model. Fine-tuning with ResNet-101 as the backbone architecture; SGD as the optimizer; steps as 1000; and learning rate as 0.001. $^a$ Starting weights mentions the pre-trained model weights for incremental training.
    }
\end{table}
\begin{table}[h!]
    \caption{Performance measure of the models shown in Table~\ref{tab:Phase2param}}
    \label{tab:Phase2M1to17Eval}
    {\scriptsize
        \centering
        \begin{tabular}{p{2cm} p{1.5cm} p{1.5cm} p{2cm}  p{2cm} p{2cm}}
            \toprule
            Model Ref No.        & AP$^{50}$                 & AP$^{75}$                 & AP$^{90}$           & mAP$^{50}$ Train       & mAP$^a$ Val  \\
            \midrule
            M1                   & 75.60               & 72.89               & 13.90         &  {96.60}  & 57.00 \\&\\
            M2                & 71.19               & 63.01               & 24.93         & 98.70            & 54.53    \\
            M3                 & 70.29               & 62.69               & 35.77         & 98.96            & 55.36                   \cr  \\
            M4                & 68.86               & 63.24               & 31.28         & 98.86            & 54.32                     \\
            M5                & 66.66               & 61.46               & 37.18         & 99.08            & 53.02                     \\
            M6               & 70.14               & 64.46               & 35.08         & 99.12            & 55.29                    \\ 
            M7                & 69.93               & 64.22               & 35.04         & 99.11            & 54.87                  \cr    \\
           
            M8               & 68.87               & 63.46               & 30.37         & 99.06            & 54.37                     \\
            M9    & 68.62               & 63.36               & 30.90         & 98.91            & 54.16                 \cr    \\
            M10             & 70.72               & 64.36               & 31.29         & 98.91            & 56.08                     \\
           \textbf{M11}       & {74.59}              & 65.31               & 41.24         & 99.13            & \textbf{59.36}            \\
            M12        & 70.26               & 64.29               & 47.38         & 99.09            & 56.21                     \\
            M13         & 70.61               & 66.50               & 49.20         & 99.11            & 57.83       \cr\\
            M14         & 70.57               & 64.08               & 40.66         & 98.96            & 56.35                      \\
            M15         & 70.40               & 63.00               & 40.77         & 99.11            & 55.54     \cr            
            \bottomrule
        \end{tabular}
    }
    {\scriptsize
    \newline
         $^a$mAP{[}0.5:0.95:0.05{]-is mean of AP of IoU threshold range from 0.5 to 0.95 with 0.05 step size }
    }
\end{table}
\begin{table}[h!]
    \centering
    \caption{Final model - M11 fine-tuning}
    \begin{tabular}{lrrr}
        \toprule
        \label{tab:fine}
        \textbf{Detection Minimum Confidence} & 50\% & 70\% & \ 90\% \\
        \midrule
        mAP\_val & 76.13 & 75.14  & \textbf{74.59} \\
        mAP\_val [0.5:0.5:0.95]: (after tuning) & \textbf{59.80} & \textbf{59.52} & 59.36 \\
        \midrule
        mAP Increase (in \%)  & \ $+0.44$\% & \ $+0.16$\% & --\\         
        \bottomrule
    \end{tabular}
\end{table}

Table~\ref{tab:fine} shows the tuning model M11's hyper-parameter \textit{detection minimum confidence} to 0.5 (50\%) and 0.7 (70\%). This tuning achieves a marginal improvement in the mAP\_val [0.5:0.05:0.95] of 0.44\% and 0.16\% over the \textit{detection minimum confidence} of 0.9 (90\%). 
Overall, it is evident from the results that with a well-planned fine-tuning approach, the model performance can be leveraged to achieve better results with a limited dataset.

\section{Conclusions}
\label{sec:con}
Mask  R-CNN, one of the seminal architectures for instance segmentation studied  for its application in segmentation of plastic waste bottles. The  challenges  such as the non-availability of a comprehensive dataset and resources for training the model were addressed by adopting a transfer learning scheme and data augmentation technique. The experiments conducted in different phases by fine-tuning the pre-trained model with a varied parameter setting showed a noticeable  improvement in the performance.  The models trained initially with \textit{head layers} on the pre-trained model and later fine-tuned with \textit{select layers} training with and without augmentation. 

In the transfer learning scheme using Mask R-CNN, applied on waste bottle instance segmentation of images, we observe that the initial \textit{head layers} training with ResNet-101 as backbone network with incremental fine-tuning achieved an AP$^{50}$ of 74.6 precision. Further, the model evaluated with mean average precision (mAP) for instance segmentation of IoU threshold range [0.5:0.95:0.5] had achieved a 59.4 mAP. Test of images and video data produced a qualitatively noticeable good performance, in instance segmentation.

In our future work, we aim to improve training dataset availability using generative adversarial networks and investigate the dense agglomerate of bottle instances, where the model ignores the object segmentation because of the non-max suppression and other configuration thresholds.


%
%
\bibliographystyle{splncs04}
\bibliography{his_pj_rv_vo_2020}
\end{document}